\documentclass[10pt,twocolumn,letterpaper]{article}

\usepackage{iccv}
\usepackage{times}
\usepackage{epsfig}
\usepackage{graphicx}
\usepackage{amsmath}
\usepackage{amssymb}
\usepackage{paralist}
\usepackage{booktabs}

\usepackage[breaklinks=true,bookmarks=false]{hyperref}

\iccvfinalcopy 

\def\iccvPaperID{0002} 

\ificcvfinal\fi

\newcommand{\xhdr}[1]{\vspace{3pt}\noindent\textbf{#1}}

\begin{document}

\title{\parbox{0.03\textwidth}{\includegraphics[width=\linewidth]{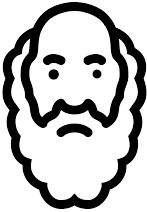}} Socratis: Are large multimodal models emotionally aware?}

\author{\textsuperscript{1*}Katherine Deng, \hspace{1pt}\textsuperscript{1*}Arijit Ray,
\hspace{1pt}\textsuperscript{1*}Reuben Tan,
  \hspace{1pt} \textsuperscript{3} Saadia Gabriel,
  \hspace{1pt} \textsuperscript{1}Bryan A. Plummer,
  \hspace{1pt} \textsuperscript{1,2}Kate Saenko \\
\textsuperscript{1}Boston University, \textsuperscript{2}Meta AI (FAIR), \textsuperscript{3}MIT\\
{\tt \small \{kdeng, array, rxtan, bplum, saenko\}@bu.edu}, {\tt \small \{skgabrie\}@mit.edu} \\
\footnotesize{$^*$Equal Contribution}} 

\maketitle

\begin{abstract}
Existing emotion prediction benchmarks contain coarse emotion labels which do not consider the diversity of emotions that an image and text can elicit in humans due to various reasons. 
Learning diverse reactions to multimodal content is important as intelligent machines take a central role in generating and delivering content to society. 
To address this gap, we propose Socratis, a \underline{soc}ietal \underline{r}e\underline{a}c\underline{ti}on\underline{s} benchmark, where each image-caption (IC) pair is annotated with multiple emotions and the reasons for feeling them. Socratis contains 18K free-form reactions for 980 emotions on 2075 image-caption pairs from 5 widely-read news and image-caption (IC) datasets.
We benchmark the capability of state-of-the-art multimodal large language models to generate the reasons for feeling an emotion given an IC pair. Based on a preliminary human study, we observe that humans prefer human-written reasons over 2 times more often than machine-generated ones. This shows our task is harder than standard generation tasks because it starkly contrasts recent findings where humans cannot tell apart machine vs human-written news articles, for instance. We further see that current captioning metrics based on large vision-language models also fail to correlate with human preferences. We hope that these findings and our benchmark will inspire further research on training emotionally aware models. Our dataset can be found at \url{https://kdeng55.github.io/socratis-website/}.

 
\end{abstract}
\section{Introduction}
\begin{figure}[t]
\centering
\includegraphics[width=0.48\textwidth]{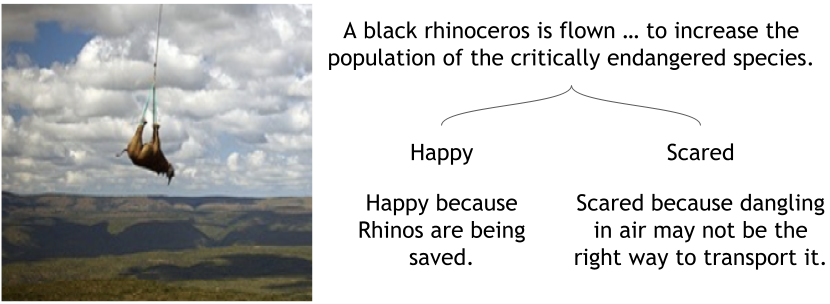}
\caption{\textbf{Socratis benchmark.} The same image and caption pair can evoke different emotions and reactions. We release a benchmark dataset of diverse human-annotated emotions and reactions to images and captions. We show that current state-of-the-art language models and metrics fail to capture the nuance of this task.}
\label{fig:teaserfig}
\end{figure}

A crucial prerequisite for effective communication and collaboration is the ability to possess emotional awareness \cite{goleman1996emotional}.
As intelligent machines become increasingly prevalent ranging from generative content creation \cite{rombach2022high} to collaborative \cite{openai2023gpt4} and embodied AI \cite{puig2020watch}, they need to possess emotional awareness for effective communication, and greater trust and acceptance \cite{goleman1996emotional}.  
Emotionally unaware messaging undermine efforts to inform people of global crises \cite{ding2011support}, spread political division \cite{vidgen2020detecting}, and fail to engage people for the correct social causes \cite{divakaran2021user}.
Existing work on emotion prediction oversimplifies the problem by categorizing emotions into coarse buckets \cite{panda2018contemplating, yang2023emoset, dalmia2016columbia}, ignoring the nuance that the same content can elicit various shades of emotional reactions for various reasons. 
For instance, as shown in Figure \ref{fig:teaserfig}, an image and caption can generate two conflicting emotions with valid reasons. 
Learning this diversity of reactions is crucial to tailor a machine's interactions to individual emotional states.

To encourage further research on emotionally aware AI, we propose the SOCRATIS benchmark - a dataset of detailed diverse reactions written by humans on images and captions. SOCRATIS includes 980 shades of emotions and 18K free-form reactions written by humans on 2075 image and caption pairs collected from 5 existing news and image-caption datasets. Given an image, a news caption and an emotion word, our task is to generate a ``reaction'' caption that explains why the image and caption may elicit the specified emotion in a person.

Unlike related benchmarks that focus on reactions to artistic images \cite{mohamed2022artelingo, achlioptas2021artemis}, niche topics such as gun violence \cite{reardon-etal-2022-bu}, emotionally stylized captions \cite{chen2021nice}, or on morality \cite{jiang2021can}, we focus on reasons why humans might feel various emotions on real-life images and captions from widely used news and image-caption datasets. Our task has higher practical relevance since models that fare well on this benchmark can be used to create more effective and inclusive messaging by news agencies and social workers. For instance, as shown in Figure \ref{fig:teaserfig}, a writer can look at why someone may be ``scared'' and update the content to reflect that transporting the rhinoceros by helicopter is safe in this context to effectively mitigate the doubt. While a similar benchmark \cite{gabriel2021misinfo} focuses on reactions to reading a text caption,
we focus reactions to both images and text since content on the web is largely multimodal.

Using Socratis as a testbed, we evaluate a state-of-the-art multimodal language model \cite{li2023blip2} and check if commonly used metrics in language-generation evaluation can distinguish between good ``reactions'' and poor ones. We generate ``reactions'' given an image, caption, and emotion and ask human raters to blindly choose between the machine generation and the human annotation. 

Our results show a stark gap in current capabilities for this task. 
Humans prefer human-written reactions two times more often than machine-generated ones. This starkly contrasts recent findings where humans fare poorly to tell the difference between machine-generated and AI-generated news articles \cite{zellers2019defending} or images \cite{rombach2022high}. This illustrates that while large generative models may be good at producing believable articles or images, it lacks the nuance required for emotional awareness. 
Furthermore, when we separate the generations into two groups - good generations as rated by humans (when humans couldn't tell the difference between human-written and machine-generated) and poor generations (when humans picked human-written over machine-generated), we observe a negligible difference in the scoring by 
commonly used metrics like BART and CLIP score, which are also based on large language and vision models. This illustrates the difficulty of the problem since we need better emotionally aware models to make better metrics.  

Hence, we hope these initial results spark further research and discussion into improving the emotional awareness of large language models. 
Adding to the recent discussion that large language models seem to lack rich social intelligence \cite{sap-etal-2022-neural} and theory of mind \cite{sclar-etal-2023-minding}, our benchmark shows that they lack nuanced emotional awareness as well.

\section{SOCRATIS Dataset}
\begin{figure*}[t]
\centering
\includegraphics[width=0.82\textwidth]{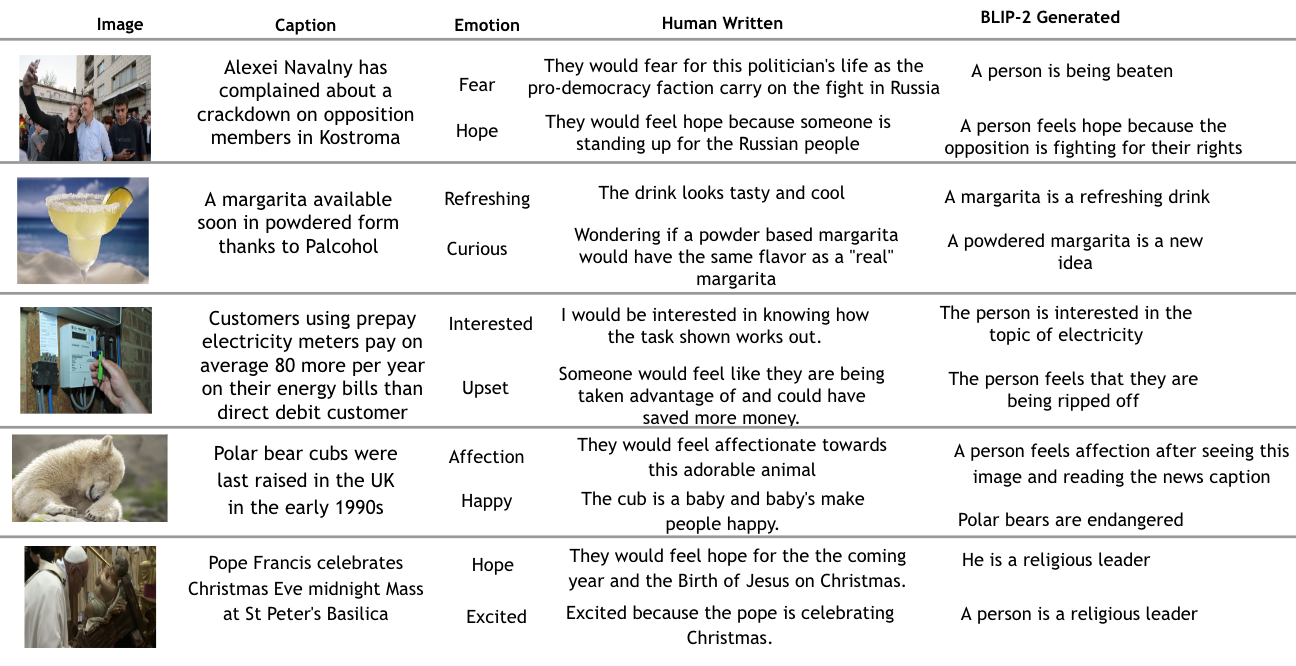}
\caption{\textbf{Qualitative examples from our Socratis dataset and a state-of-the-art multimodal model, BLIP-v2 generations.}}
\label{fig:qual_examples}
\end{figure*}

We propose a benchmark for evaluating the emotional awareness of vision-language models. Specifically, our task is to predict the reaction a human may have of a certain emotion while looking at an image and caption (IC) pair.
To this end, we collect a dataset showing human workers an IC pair and ask them to write the emotion words they feel and the reasons for feeling them. We interchangeably call these reasons as ``reactions'' the humans feel of a certain emotion.
We have 18,378 annotated reactions for 980 emotions to 2075 IC pairs rated by an average of 8 independent workers per IC pair. The most common emotion words follow that of standard emotion datasets \cite{panda2018contemplating, dalmia2016columbia} - happy, sad, excited, angry. We also have a variety of more subtle emotion words such as inspire, hopeful, and nostalgic that feature among the top 30 emotion words.  
However, the novelty of our dataset isn't the variety of emotion words, but the free-form reasons for feeling the different emotions for the same IC pair. Some examples are shown in Figure \ref{fig:qual_examples}.
We will make our dataset publicly available. 


\subsection{Data Annotation}

\xhdr{Image-Caption pair collection}
We collect image-caption pairs from the Visual News \cite{liu2020visual} and the Conceptual Captions \cite{sharma2018conceptual} dataset. We first randomly sample 1800 images from the Visual News Dataset \cite{liu2020visual}, which consists of an image and a news headline caption from 4 news datasets - BBC, USA-Today, Guardian and The Washington Post. We specifically choose news datasets since news headlines and visuals usually elicit stronger reactions from humans than generic stock images and are of more practical use. Additionally, to also understand how people react to generic IC pairs, we sample images and captions from a widely used image-captioning dataset, Conceptual Captions \cite{sharma2018conceptual}. IC pairs that explicitly convey a certain emotion (eg, stock photo of someone smiling) are less interesting due to a lack of ambiguity from diverse people. Hence, to sample IC pairs that are likely to elicit diverse emotional reactions, we choose 500 samples where the emotion of the image does not match the emotion of the caption. The emotion of the image is predicted by a CLIP \cite{radford2021learning} model, fine-tuned on the WebEmo \cite{panda2018contemplating} dataset, and a T5 model \cite{raffel2020exploring} predicts the text emotions. 

\xhdr{Human Reaction Annotation}
We show users the IC pairs we collect as described above. 
We then ask them to write up to three emotions a person is likely to feel while viewing the IC pair, along with a reason for why they might feel each of the emotions they entered.
We collect our dataset on Amazon Mechanical Turk. All data is cleared of personally identifiable information and only aggregate statistics of model training are reported in the paper.
We incentivize the annotators by awarding them a small bonus if they can match the most popular emotion word for an IC pair. 
This encourages workers not to enter noisy reactions (or extremely uncommon or made-up words) since they are not likely to match other workers' responses. 
To control the quality of responses,  we choose workers with a greater than 98\% approval rate (on at least 50 Human Intelligence Tasks or HIT's). 
We also restrict the geographical location to the US or UK since the images and captions are sourced from news articles from these two countries. 
We collect 10 independent annotations for each image-caption pair to get a representative, diverse set of reactions. 



\subsection{Annotation Quality}
\noindent We compute some heuristic automatic measures to judge the quality of the reactions.

\xhdr{Emotion-reaction match}
We check how often the entered emotion word matches the emotion of the reaction. For instance, the emotion of ``sad'' reaction should also be ``sad''.   
To compute the sentiment, we use a T5 sentiment model \cite{raffel2020exploring}, which predicts positive or negative sentiment. 
We observe an $87.15\%$ accuracy of sentiment match between the emotion word and the reaction. 
We manually check some examples where the sentiments of reaction and emotion words mismatch and find that the T5 model \cite{raffel2020exploring} may be noisy. For instance, for the emotion word ``hungry", the reaction annotated is ``the cocoa actually makes me want to eat something sweet.". We believe this is a reasonable reaction for ``hungry''. However, the sentiment predicted for ``hungry'' is ``negative'', whereas the same for the reaction is ``positive", resulting in a mismatch. 
This further illustrates that coarse emotion buckets like positive and negative don't capture the nuances of reactions of what people feel. 

\xhdr{Agreement on reactions}
To judge the agreement of humans over the reactions for a given image, caption, and emotion, we further compute the BART \cite{yuan2021bartscore} scores of the reactions for the same (I,C,emotion) tuple. We contrast this with the BART score \cite{yuan2021bartscore} of 1000 randomly sampled reaction pairs from different IC-pairs and emotions. We note that the BART scores between reactions for the same emotion for an IC pair are higher $78\%$ of the time than that of the random pairs. 

\section{Approach}
To understand the emotional awareness of large multimodal models, we benchmark the capability of a state-of-the-art vision-language model on our proposed benchmark without further fine-tuning. 
Specifically, we use the FLAN-T5 variant of BLIP-2 \cite{li2023blip2} from Huggingface \cite{wolf2019huggingface}. 
Our task is to predict the reason of feeling a certain emotion given an image-caption pair. 
For a given image, caption text and emotion word, we prompt BLIP-2 with the image and a query using the following template:
\texttt{Question: Why does a person feel \{emotion\} after seeing this image and reading the news caption `\{caption\}'? Answer:}

\noindent We input this prompt along with the image to the BLIP-2 model, and use a greedy approach to generate the response following the standard procedure outlined on the Huggingface \cite{wolf2019huggingface} model page. 
Given an image $i$, caption $c$ and emotion $e$, we can formulate the likelihood of the generated reaction text $t$ as:
\begin{equation}
    P(t|i, c, e) = \Pi_{j=1}^{n}P(t_{<j} | i, c, e)
\end{equation}
where $n$ denotes the number of tokens in the generated text.
\section{Experiments}
Our goal is to evaluate whether large multimodal models are emotionally aware to generate plausible reasons for why humans might feel a certain reaction for a given image and text.
A pragmatic formulation of this evaluation is whether a human finds it hard to distinguish a machine generated reason from a human-written reason.
Hence, we first conduct a human evaluation on the multimodal BLIP-2 generated reasons to see how often humans prefer machine-generated reasons, human-written reasons, or both. 

\xhdr{Human Evaluation}
To understand human preference, we conduct a preliminary human study on 500 randomly sampled data points from our dataset.
We show an image, caption, emotion tuple to a user. We then show two choices of reactions - human-written and machine-generated. 
The user is unaware of which of the reactions is machine-generated or human-written. 
We ask 3 independent workers to choose the best reaction for the given tuple of image, caption and emotion. They also have the choice of either choosing that both reactions are reasonable or that neither is reasonable. 
Hence, there are four choices for each image-caption-emotion tuple: 
We split the human study image-caption-emotion tuples into three groups:
\begin{compactenum}[--]
\item \textbf{Machine-better}: machine was picked over human, indicating that the machine-generations are good. 
\item \textbf{Both-Good}: both human-written and machine-generated reactions are equally valid.
\item \textbf{Both-Bad}: We discard these examples from further study since the data annotations are likely noisy. 
\end{compactenum}

\xhdr{Metrics}
Since human evaluations are slow and expensive, we aim to determine if commonly used captioning metrics can be used to judge good reaction generations from poor ones to speed up research. We define a good generation as one which is aligned to or better than human preferences. Hence, a machine generation is good if a human cannot tell the difference from a human-written one, or if a human prefers it over a human-written one. Hence, a good metric should score such cases (machine-preferred and both-good) higher than the cases where a human-written reason was preferred over a machine-generated one (human-preferred). We compute three commonly used metrics. First, we compute \textbf{BART Score} between the machine-generated reaction and the human-generated reference to measure human-likeness of generations as described in \cite{yuan2021bartscore}. Next, we compute \textbf{CLIP-Score} and \textbf{RefCLIP-Score} \cite{hessel2021clipscore} to see if image-relevance plays a major factor in distinguishing good generations from poor ones. We use a prompt like ``Human feels \{emotion\}, when seeing this image with \{caption\} because \{explanation\}'' and compute the cosine similarity to the image as described in \cite{hessel2021clipscore}. 

\begin{table}[]
\caption{Number of times majority ($\frac{2}{3}$) humans prefer human-written reactions, BLIP-2 (machine) reactions, or both out of 382 examples.}
\centering
\begin{tabular}{@{}llll@{}}
\toprule
\small{Machine-better} & \small{Human-better} & \small{Both-Good} & \small{Both-Bad} \\ \midrule
\hspace{.5cm}91     &   \hspace{.3cm}233    & \hspace{.6cm}47 & \hspace{.5cm}11     \\ \bottomrule
\end{tabular}
\label{tab:human_pref}
\end{table}

\begin{table}[]
\caption{Evaluations with BART and CLIP-Score on subsets where humans prefer human-written reactions, BLIP-2 (machine) reactions, or both. We want machine-better or both-good generations scored higher than human-better.}
\centering
\begin{tabular}{@{}lccc@{}}
\toprule
& \small{Machine-better} & \small{Human-better} & \small{Both-Good} \\ \midrule 
\small{BART} &  -5.48    &   -5.42    &   -5.54 \\ 
\small{CLIP-Score}   &   0.74       &    0.76    &  0.75 \\ 
\small{RefCLIP}   &    0.42     &   0.43    &  0.42\\ \bottomrule 
\end{tabular}
\label{tab:metric_eval}
\end{table}

\begin{table}[]
\caption{Multimodal vs a text-only model on relevance of generation to the image}
\centering
\begin{tabular}{@{}llll@{}}
\toprule
        &  \small{CLIP $\uparrow$} & \small{Ref-CLIP $\uparrow$} \\ \midrule
\small{BLIP-2}  &  \textbf{0.75}   &  0.42      \\
\small{FLAN-T5} &   0.73   &  0.42     \\ \bottomrule
\end{tabular}
\label{tab:im_vs_language}
\end{table}

\section{Results}

\xhdr{Humans prefer human-written reactions to machine generations.} In Table~\ref{tab:human_pref}, we observe that workers pick human reactions over two times more often than machine-generated ones (233 vs 91). This suggests that state-of-the-art large vision-and-language models are still limited at extrapolating contextual information beyond simply correlating the visual concepts in the image with relevant words.

\xhdr{Current captioning metrics cannot distinguish between good and bad reactions} In Table~\ref{tab:metric_eval}, we see that BART scores do not follow human preferences. The scores of the machine generations for when the machine was rated better or equally good (both-good) by humans are not higher than the scores when the machine generations are poor (human-better). CLIP and RefCLIP also do not seem to differ across the three sets. This suggests that visual similarity may also not be important in distinguishing good from poor reactions. Further investigation is required to check if we can train a custom BART metric based on a few examples in our dataset.

\xhdr{Multimodal models are slightly more image-relevant}
We also check the performance of a language-only model compared to the multimodal BLIP-2 by using only the langauge model in BLIP2, which is a FLAN-T5 model. In Table~\ref{tab:im_vs_language}, we see that the relevance to the image is understandably higher (acc to CLIP scores) for BLIP-2. However, based on results in Table~\ref{tab:metric_eval}, this doesn't necessarily mean that the generations are more preferred by humans. 

\xhdr{Discussion and Conclusion} We propose the Socratis benchmark to generate reactions for why images and captions elicit certain emotions in humans. Our initial experiments indicate that a state-of-the-art vision-language model is unable to extrapolate the required information to generate reasonable reactions. However, further research is required to investigate biases that exist in our dataset and how these are perpetuated in the models that are benchmarked on it. Further investigation is also required to see if quick fixes like changing the generation strategy, adapting a few layers, or in-context learning (showing examples in prompts) can make these models more emotionally aware. We hope our Socratis benchmark will encourage future research on training and evaluating emotionally aware AI algorithms.

\xhdr{Acknowledgements}: We are thankful to Praneeth Chandra Bogineni for valuable discussions in the initial phase of the project. This material is based upon work supported, in part, by DARPA under agreement number HR00112020054. The findings in the paper do not reflect the opinions of the US Government or DARPA. 

{\small
\bibliographystyle{ieee_fullname}
\bibliography{egbib}
}

\end{document}